\documentclass[11pt]{article}

\usepackage[preprint]{acl}

\usepackage{times}
\usepackage{latexsym}

\usepackage[T1]{fontenc}

\usepackage[utf8]{inputenc}

\usepackage{microtype}

\usepackage{inconsolata}

\usepackage{graphicx}
\usepackage{xcolor}
\usepackage{fontawesome}
\usepackage[table]{xcolor} 

\usepackage{makecell}
\newcommand*\rot{\rotatebox{90}}

\title{PL-MTEB: Polish Massive Text Embedding Benchmark}

\author{Rafał Poświata, Sławomir Dadas, Michał Perełkiewicz\\
  National Information Processing Institute \\
  al. Niepodległości 188b, 00-608 Warsaw, Poland \\
 {\faEnvelopeO} \! \texttt{rposwiata@opi.org.pl} \\} 

\begin{document}
\maketitle
\begin{abstract}
In this paper, we introduce the Polish Massive Text Embedding Benchmark (PL-MTEB), a comprehensive benchmark for text embeddings in the Polish language. PL-MTEB comprises 30 diverse NLP tasks across five categories: classification, clustering, pair classification, information retrieval, and semantic text similarity. Within the scope of this work, we added 12 new Polish-language tasks to MTEB based on existing datasets and prepared two new datasets used to create four clustering tasks. We evaluated 30 publicly available text embedding models, including Polish and multilingual models. We analyzed the results in detail for specific task types and model sizes. We made the prepared datasets, the source code for evaluation, and the obtained results available to the public at \url{https://github.com/rafalposwiata/pl-mteb}.
\end{abstract}

\section{Introduction}

Text embeddings are used in many NLP tasks, including document clustering \citep{Aggarwal2012}, semantic search \citep{10.1145/3394486.3403305}, question answering \citep{karpukhin-etal-2020-dense} or classification \citep{muennighoff-etal-2023-mteb}. In many cases, they are fundamental elements of the created systems and significantly impact their performance. Therefore, it is important to select the appropriate embedding model based on the results of its evaluation. Most often, evaluation is conducted on individual tasks using a limited set of datasets, leaving the open question of how such embedding models would work for other tasks. To solve this problem, \cite{muennighoff-etal-2023-mteb} created a Massive Text Embedding Benchmark (MTEB). MTEB provides a simple and clear way to examine how the model behaves for different types of tasks. Most of the tasks in MTEB were based on English-language datasets, and only a few were multilingual, making it impossible to do a good comparison of models for languages other than English. Therefore, extensions to MTEB with language-specific task sets have begun to appear, among which are C-MTEB \citep{c-mteb} for Chinese, MTEB for French \citep{ciancone2024extending}, FaMTEB \cite{zinvandi-etal-2025-famteb} for Persian, MTEB-NL \cite{mteb-nl} for Dutch, ruMTEB \citep{snegirev-etal-2025-russian} for Russian, VN-MTEB \citep{pham2025vnmtebvietnamesemassivetext} for Vietnamese, TR-MTEB \citep{baysan-gungor-2025-tr} for Turkish, SEB \citep{enevoldsen2024scandinavian} for Scandinavian languages (Danish, Norwegian, Swedish), ArabicMTEB \citep{bhatia-etal-2025-swan} for Arabic languages and AfriMTEB \citep{uemura2025afrimtebafrie5benchmarkingadapting} for African languages. In addition, the Massive Multilingual Text Embedding Benchmark (MMTEB) \citep{mmteb} initiative was launched, a community-driven, large-scale expansion of MTEB, covering more than 500 quality-controlled evaluation tasks in 250+ languages. In this work, we follow this path by introducing PL-MTEB (Polish Massive Text Embedding Benchmark), a comprehensive benchmark for text embeddings for Polish. Below we highlight the main contributions of this work:

\begin{itemize}
\item Introduction of PL-MTEB: a comprehensive benchmark consisting of 30 tasks from 5 groups (classification, clustering, pair classification, retrieval, and semantic textual similarity), designed to evaluate text embeddings for the Polish language.
\item Extension of MTEB with 12 new tasks based on existing Polish datasets.
\item Preparation of two new datasets: PLSC (Polish Library of Science Corpus) and Wikinews-PL. The collections were used as a basis for proposing four new tasks for clustering.
\item Evaluation of 30 models (12 for Polish and 18 multilingual) with collection of results.
\item Integration with MTEB and public release of source code, all experimental results and prepared datasets.
\end{itemize}

\section{Related work}

\subsection{Benchmarks}
\label{sec:benchmarks}
GLUE \citep{wang-etal-2018-glue} or SuperGLUE \citep{10.5555/3454287.3454581} are well-known benchmarks for tracking NLP progress. They are mainly designed to compare natural language understanding systems. However, they are unsuitable for evaluating text embeddings, so dedicated benchmarks like SentEval \citep{conneau-kiela-2018-senteval} or BEIR \citep{thakur2021beir} have emerged. MTEB \citep{muennighoff-etal-2023-mteb} incorporates the above benchmarks, creating an accessible evaluation framework. In the following years, extensions to MTEB were introduced, covering various languages, as we mentioned at the beginning of this paper. 

For Polish, benchmarks similar to (Super)GLUE include KLEJ \citep{rybak-etal-2020-klej} and LEPISZCZE \citep{NEURIPS2022_890b206e}. Previously, in most cases, text embedding evaluation for the Polish language was performed on individual tasks. \cite{krasnowska-kieras-wroblewska-2019-empirical} evaluated text embeddings on a single dataset for textual relatedness. 
\cite{dadas-etal-2020-evaluation}, in their evaluation, used 3 task types (classification, textual entailment, and semantic relatedness), where only classification consisted of more than one task.
\cite{dadas-para} extended this evaluation by adding 3 more tasks, one of each type. In the field of information retrieval, two broader benchmarks for Polish have emerged recently. The first is BEIR-PL \citep{wojtasik2023beirpl}, which is the Polish equivalent of BEIR \citep{thakur2021beir}. The second is PIRB \citep{dadas-etal-2024-pirb-comprehensive}, a large benchmark consisting of 41 tasks. 

\subsection{Embedding Models}
A few years ago, the standard method for creating text embeddings was to compute arithmetic or weighted averages of the word vectors in a text. These vectors were obtained using word embedding models such as Word2Vec \citep{word2vec_nips, word2vec_iclr}, GloVe \citep{pennington-etal-2014-glove}, or FastText \citep{bojanowski-etal-2017-enriching}. The main disadvantage of these methods was the lack of context awareness.
The emergence of the Transformer \citep{NIPS2017_3f5ee243} architecture, introducing context awareness through the use of the self-attention mechanism, forms the foundation of most recent embedding models. \cite{reimers-gurevych-2019-sentence} have shown that additional fine-tuning of a network composed of two transformer models leads to a model that produces high-quality sentence embeddings. Further development of the field is mainly models that use contrastive loss objective, among which we can include: SimCSE \citep{gao-etal-2021-simcse}, TSDAE \citep{wang-etal-2021-tsdae-using}, GTR \citep{ni-etal-2022-large}, SGPT \citep{muennighoff2022sgpt}, E5 \citep{wang2022text}, or BGE \citep{c-mteb}. Although most of the models were designed for English, some multilingual models included Polish. Among these models, we can highlight multilingual E5 \citep{wang2022text} and Arctic-Embed 2.0 \citep{yu2024arcticembed20multilingualretrieval} models.
With the rapid development of large language models, new text embedding methods based on them are now becoming available. Among them, the following can be distinguished:
Qwen3-Embedding \citep{qwen3embedding}, BGE-Gemma2 \citep{c-mteb, bge-m3} or KaLM-Embedding \citep{kalm} models series. Models developed specifically for the Polish language were mostly created using a multilingual knowledge distillation technique \citep{reimers-gurevych-2020-making} and Polish-English bilingual corpora. Among these models are Polish SBERT \citep{dadas-para}, the MMLW \citep{dadas-etal-2024-pirb-comprehensive} models series, and Stella-PL \citep{dadas-etal-2024-pirb-comprehensive}. 

\section{PL-MTEB Benchmark}

\begin{figure*}[h]
\begin{center}
\includegraphics[scale=1]{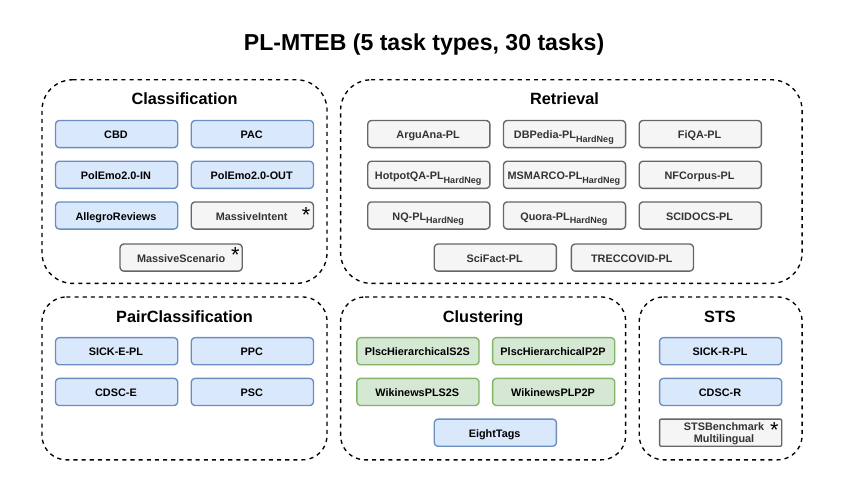} 
\caption{An overview of tasks included in PL-MTEB. The tasks with \colorbox[HTML]{F5F5F5}{gray} background are tasks in Polish that are already in MTEB (those marked with an * are multilingual tasks from which we have selected Polish subtasks). The tasks marked in \colorbox[HTML]{DAE8FC}{blue} are tasks prepared in this work based on existing datasets. The \colorbox[HTML]{D5E8D4}{green} tasks were prepared on the basis of newly created datasets.}
\label{fig:pl-mteb}
\end{center}
\end{figure*}

\subsection{Task Types and Metrics}

The benchmark consists of the following five task types:

\paragraph{Classification} The classification task is to predict a label from an input embedding using a previously trained logistic regression classifier. A small subset of examples (8 per class) is randomly selected from the entire training set, so that results are less influenced by the training data and more by the encoding method. The process is repeated 10 times, each time with a different set of training examples. The reported results are the average of all these experiments. The metrics used in this task are accuracy, F1-score, precision, and recall, with the last three calculated in both macro and weighted versions. The accuracy is used as the main metric.

\paragraph{Clustering} Given a set of sentences or paragraphs, clustering aims to group them into meaningful clusters. A mini-batch k-means model with a batch size of 512 and k equal to the number of distinct labels is trained on the embedded texts. This process is repeated 10 times, and the result is the average of the experiments. The model is scored using v-measure \citep{rosenberg-hirschberg-2007-v}. In hierarchical clustering, evaluation is performed at each level, and the reported result is the average v-measure across all levels.

\paragraph{Pair Classification} Having a pair of embedded texts, predict their relationship as a binary label based on their similarity. The calculated measures are precision, average precision, recall, accuracy, and F1-score, based on cosine similarity, dot product, Euclidean distance, and Manhattan distance. The average precision score based on cosine similarity is the main metric.

\paragraph{Retrieval} The retrieval task is presented with a corpus, queries, and a mapping for each query to relevant documents from the corpus. The provided model is used to embed all queries and all corpus documents. The goal is to find relevant documents based on the query. Various metrics are used to measure retrieval performance, including MAP@N, nDCG@N, MRR@N, precision@N, and recall@N, where N is from \{1, 3, 5, 10, 20, 100, 1000\}. The nDCG@10 serves as the main metric.

\paragraph{Semantic Textual Similarity (STS)} Given a pair of sentences, the goal is to measure their correlation using the similarity score between their embeddings. Spearman and Pearson correlation coefficients are computed based on cosine similarity, Euclidean, and Manhattan distances. Spearman correlation based on cosine similarity is the main metric.

\subsection{Tasks}
\label{subsec:tasks}

Figure \ref{fig:pl-mteb} provides an overview of the tasks available in PL-MTEB. The tasks have been categorized by origin. The first group (\colorbox[HTML]{F5F5F5}{gray}) contains tasks in Polish or multilingual tasks containing a subtask in Polish, added to MTEB by other contributors. These are mainly retrieval tasks from the BEIR-PL \citep{wojtasik2023beirpl} benchmark. Tasks with the HardNeg suffix refer to cases where the original corpus of passages has been reduced and restricted to relevant passages and a specified number of hard negatives. Limiting the number of passages significantly speeds up the evaluation process and was proposed in MMTEB \citep{mmteb}. The second group (\colorbox[HTML]{DAE8FC}{blue}) contains tasks we added based on existing datasets. When selecting the datasets for these tasks, we focused primarily on their public availability and the method used to prepare them, which involved manual annotation and verification by native Polish speakers. Most of the datasets we adopted came from the works described in subsection \ref{sec:benchmarks}, with the majority from the KLEJ \citep{rybak-etal-2020-klej} benchmark. The third group (\colorbox[HTML]{D5E8D4}{green}) contains tasks we added based on newly created datasets. When compiling tasks from the two previous groups, we noticed a significant underrepresentation of clustering tasks; therefore, to fill this gap, we prepared two new datasets on which we based four clustering tasks.   
In the following subsections, we discuss these new datasets and the data quality verification process we conducted for all the tasks we added.

\begin{table*}[h]
    \scriptsize
    \centering
    \begin{tabular}{lllll}
\hline
 \bf Task                        & \bf Reference   & \bf Test samples   & \bf Domains   & \bf Dataset Licence                              \\
\hline
\multicolumn{5}{l}{\textbf{Classification}} \\
 \hline
 CBD                         & \citet{ptaszynski2019results}         & 999            & Written, Social                                          & BSD-3-CLAUSE                         \\
 PolEmo2,0-IN                     & \citet{kocon-etal-2019-multi}        & 722            & Written, Social                                          & CC-BY-SA-4.0                         \\
 PolEmo2.0-OUT                    & \citet{kocon-etal-2019-multi}        & 493            & Written, Social                                          & CC-BY-SA-4.0                         \\
 AllegroReviews              & \citet{rybak-etal-2020-klej}         & 983            & Reviews                                                  & CC-BY-SA-4.0                         \\
 PAC                         & \citet{NEURIPS2022_890b206e}         & 3,395            & Legal, Written                                           & CC-BY-NC-SA-4.0                      \\
 MassiveIntent               & \citet{fitzgerald2022massive}         & 2,974            & Spoken                                                   & APACHE-2.0                           \\
 MassiveScenario             & \citet{fitzgerald2022massive}          & 2,974            & Spoken                                                   & APACHE-2.0                           \\
\hline
\multicolumn{5}{l}{\textbf{Clustering}} \\
 \hline
 EightTags                   & \citet{dadas-etal-2020-evaluation}         & 2,048           & Social, Written                                          & GPL-3.0                              \\
 PlscHierarchicalS2S         & PL-MTEB         & 2,048            & Academic, Written                                        & CC0-1.0                              \\
 PlscHierarchicalP2P         & PL-MTEB        & 2,048            & Academic, Written                                        & CC0-1.0                              \\
 WikinewsPlS2S               & PL-MTEB         & 2,048           & News                                                     & CC-BY-4.0                            \\
 WikinewsPlP2P               & PL-MTEB         & 2,048            & News                                                     & CC-BY-4.0                            \\
 \hline
\multicolumn{5}{l}{\textbf{Pair Classification}} \\
 \hline
 SICK-E-PL                   & \citet{dadas-etal-2020-evaluation}         & 4,874            & Web, Written                                                  &              CC-BY-NC-SA-3.0                        \\
 CDSC-E                      & \citet{wroblewska2017polish}        & 998            & Web, Written                                                   & CC-BY-NC-SA-4.0                      \\
 PSC                         & \citet{ogro:kop:14:lrec}         & 1,074            & News, Written                                            & CC-BY-3.0                            \\
 \makecell[l]{PPC \\ \!}                     & \makecell[l]{\citet{dadas-para} \\ \!}          & \makecell[l]{1,000 \\ \!}             & \makecell[l]{Fiction, Non-fiction, Web, \\ Written, Spoken, Social, News} & \makecell[l]{GPL-3.0 \\ \!}                             \\
 \hline
\multicolumn{5}{l}{\textbf{Retrieval}} \\
 \hline
 ArguAna-PL               & \citet{wojtasik2023beirpl}        &  1,406 / 8,674             &  Medical, Written                                      &  CC-BY-SA-4.0                          \\
 DBPedia-PLHardNeg     & \citet{wojtasik2023beirpl, mmteb}          & 400 / 88,542            & Written, Encyclopaedic                                   & MIT                                  \\
 FiQA-PL                     & \citet{wojtasik2023beirpl}         & 648 / 57,638            & Written, Financial                                       & NOT SPECIFIED                        \\
 HotpotQA-PLHardNeg    & \citet{wojtasik2023beirpl, mmteb}         & 1,000 / 212,774            & Web, Written                                             & CC-BY-SA-4.0                         \\
 MSMARCO-PLHardNeg     & \citet{wojtasik2023beirpl, mmteb}          & 43 / 9,481            & Web, Written                                             & OWN LICENCE \\
 NFCorpus-PL                 & \citet{wojtasik2023beirpl}         & 323 /  3,633           &     Medical, Academic, Written                                                     &       NOT SPECIFIED                                 \\
 NQ-PLHardNeg         & \citet{wojtasik2023beirpl, mmteb}          & 1,000 / 184,765            &    Written, Encyclopaedic                                                     &          CC-BY-NC-SA-3.0                             \\
 Quora-PLHardNeg       & \citet{wojtasik2023beirpl, mmteb}         & 1,000 / 172,031            &    Written, Web, Blog &                                            NOT SPECIFIED                                             \\
 SCIDOCS-PL                  & \citet{wojtasik2023beirpl}         & 1,000 / 25,657            &    Academic, Written, Non-fiction                                                      &    CC-BY-SA-4.0                                   \\
 SciFact-PL                  & \citet{wojtasik2023beirpl}         & 300 / 5,183           & Academic, Medical, Written                               &               NOT SPECIFIED                       \\
 \makecell[l]{TRECCOVID-PL \\ \!}               & \makecell[l]{\citet{wojtasik2023beirpl} \\ \!}        & \makecell[l]{50 / 171,332 \\ \!}            & \makecell[l]{Academic, Medical, \\ Non-fiction, Written}                  & \makecell[l]{NOT SPECIFIED \\ \!}                       \\
 \hline
\multicolumn{5}{l}{\textbf{STS}} \\
 \hline
 SICK-R-PL                   & \citet{dadas-etal-2020-evaluation}        & 4,871            & Web, Written                                             & CC-BY-NC-SA-3.0                      \\
 CDSC-R                      & \citet{wroblewska2017polish}        & 998            & Web, Written                                             & CC-BY-NC-SA-4.0                      \\
 \makecell[l]{STSBenchmarkMultilingual \\ \!} & \makecell[l]{\citet{huggingface:dataset:stsb_multi_mt} \\ \!}         & \makecell[l]{1,379 \\ \!}            & \makecell[l]{News, Social, Web, \\ Spoken, Written}                       & \makecell[l]{NOT SPECIFIED  \\ \!}                       \\
\hline
\end{tabular}
\caption{\label{tab:tasks}
Tasks in PL-MTEB. The two numbers in the test samples column for the retrieval tasks represent the number of questions and the corpus size, respectively. The domains specify the source of the texts in each task.
}
\end{table*}

\subsubsection{New datasets}

\noindent
\textbf{PLSC} (Polish Library of Science Corpus) is a dataset based on Library of Science\footnote{\href{https://bibliotekanauki.pl/}{https://bibliotekanauki.pl/}}, an open metadata repository about scientific publications. Using the provided API, we retrieved publication metadata, including the title, abstract, journal, and assigned categories. We divided the categories into scientific fields and scientific disciplines, with each scientific discipline assigned to a specific field, creating a hierarchical relationship. In this way, each record was assigned to at least one of the 8 fields and 44 disciplines. The next step was to verify the abstract language using the langdetect\footnote{\href{https://pypi.org/project/langdetect/}{https://pypi.org/project/langdetect/}} library, as some records pertained to publications affiliated with Poland but written in other languages. We discarded records in languages other than Polish. The corpus comprises about 160K records. To prepare the tasks, we selected only those records from the collected data that were assigned to exactly one field and one discipline. We randomly limited the number of records to 200 per discipline.
This collection was used to prepare two clustering tasks: PlscHierarchicalS2S and PlscHierarchicalP2P\footnote{This is inspired by tasks from MTEB, such as ArxivS2S and ArxivP2P. S2S (sentence to sentence) and P2P (paragraph to paragraph) mean that
the sentence/paragraph is compared with another sentence/paragraph, where the paragraph is a longer fragment of text, e.g., title + abstract.}, where for the S2S task, publication titles were used, while for the P2P task, titles were combined with the abstract. The tasks were hierarchical, i.e., first there was clustering by scientific fields, then by scientific disciplines, and the results were averaged. For performance reasons, the number of records has been limited to 2,048, in accordance with MMTEB \citep{mmteb} assumptions.

\noindent
\textbf{Wikinews-PL} is a dataset of articles from the Polish version of the Wikinews portal\footnote{\href{https://pl.wikinews.org}{https://pl.wikinews.org}} . Each article is assigned to one or more categories among the following: politics, economy, disasters, culture and entertainment, science, law and crime, sports, society and technology. 
The collection we downloaded consists of 15,196 articles. To prepare the WikinewsPLS2S and WikinewsPLP2P clustering tasks, we selected only those records that are assigned to a single category. We preprocessed the text by removing timestamps appearing at the beginning of some articles. We randomly limited the number of records per category to 500, and then, as before, reduced the entire resulting dataset to 2,048. For the S2S task, article titles were used, while for the P2P task, titles were combined with the main body of the article.  

\subsubsection{Data Quality}
During task preparation, we verified data quality by adjusting the functions introduced in newer versions of the MTEB framework. First, we removed examples that were empty strings and shorter than three words. Next, we verified the labels and scores. If there were near duplicates\footnote{To detect near duplicates, texts were normalized by converting them to lowercase and removing spaces.} with different labels or with a score difference of at least 0.5, we removed them. The next step was deduplication at the split level, where we first remove exact duplicates and then near duplicates. The final step was to verify that there was no test-train leakage. As a result of this process, we obtained datasets used to prepare the PL-MTEB tasks. 

A summary of the PL-MTEB tasks is presented in Table \ref{tab:tasks}. All of the tasks are based on datasets under open licenses and are publicly available on the Hugging Face Hub\footnote{\href{https://huggingface.co/datasets}{https://huggingface.co/datasets}}. For more information about the tasks, see Appendix \ref{sec:tasks_appendix}.

\definecolor{best}{HTML}{BDD7EE}
\definecolor{second_best}{HTML}{DDEBF7}
\definecolor{good}{HTML}{DDEBF7}

\begin{table*}[h!]
\scriptsize
\centering
\begin{tabular}{l|rr|ccccc|cc}
\hline
 \makecell[l]{\bf Model name \\ \bf  / (\# tasks)}  &  \makecell[c]{\bf Model \\ \bf size} & \makecell[c]{\bf Zero \\ \bf shot} & \makecell{\bf Class. \\ \bf (7)} & \makecell{\bf Clust. \\ \bf (5)} &  \makecell{\bf PairClass. \\ \bf (4)} &  \makecell{\bf Retr. \\ \bf (11)} &  \makecell{\bf STS \\ \bf (3)} &  \makecell{\bf Avg. \\ \bf (30)}  &    \makecell{\bf Avg. \\ \bf (by type)}  \\
\hline
 \multicolumn{8}{l}{\textbf{Small models (< 150M)}} \\
 \hline
static-similarity-mrl-multilingual-v1   & 108M   &          96 & 48.17            & 30.04          & 70.41                & 24.84         & 72.01         & 41.95          & 49.09               \\
 paraphrase-multilingual-MiniLM-L12-v2   & 118M   &          93 & 51.39            & 40.68          & 83.40                & 30.40         & 78.68         & 48.91          & 56.91               \\
 multilingual-e5-small                   & 118M   &          90 & 52.64            & 43.99          & 81.70                & 46.00         & 78.41         & 55.21          & 60.55               \\
 mmlw-e5-small                           & 118M   &          90 & 60.12            & \cellcolor{good}{48.91} & 86.67                & 46.43         & 82.05         & 58.97          & 64.84               \\
 st-polish-paraphrase-from-distilroberta & 124M   &         100 & 57.71            & 42.71          & 86.96                & 36.16         & 82.63         & 53.70          & 61.23               \\
 silver-retriever-base-v1.1              & 124M   &         100 & 57.03            & 44.92          & 74.82                & 42.92         & 74.61         & 53.97          & 58.86               \\
 st-polish-paraphrase-from-mpnet         & 124M   &         100 & 57.57            & 44.53          & 87.06                & 38.33         & 82.83         & 54.80          & 62.06               \\
 mmlw-roberta-base                       & 124M   &          96 & \cellcolor{good}{62.53}   & 48.00          & \cellcolor{good}{88.16}       & \cellcolor{good}{53.6} & \cellcolor{good}{85.2} & \cellcolor{good}{62.52} & \cellcolor{good}{67.50}       \\
 distiluse-base-multilingual-cased-v2    & 135M   &          93 & 48.95            & 38.86          & 79.37                & 24.68         & 75.75         & 45.10          & 53.52               \\
 \hline
 \multicolumn{8}{l}{\textbf{Base models}} \\
 \hline
 drama-base                            & 212M   &          90 & 42.06            & 40.48         & 72.05                & 28.29         & 65.01          & 43.04          & 49.58               \\
 mmlw-e5-base                          & 278M   &          90 & 47.37            & 37.29         & 59.50                & \cellcolor{good}{53.70} & 49.02          & 49.80          & 49.38               \\
 paraphrase-multilingual-mpnet-base-v2 & 278M   &          93 & 53.23            & 41.34         & \cellcolor{good}{86.21}       & 33.33         & \cellcolor{good}{81.13} & 51.14          & 59.05               \\
 multilingual-e5-base                  & 278M   &          90 & \cellcolor{good}{55.36}   & \cellcolor{good}{44.10} & 82.08                & 47.63         & 79.13          & 56.59          & \cellcolor{good}{61.66}      \\
 snowflake-arctic-embed-m-v2.0         & 305M   &          90 & 54.01            & 43.80         & 78.37                & 52.21         & 75.60          & \cellcolor{good}{57.06} & 60.80               \\
 \hline
 \multicolumn{8}{l}{\textbf{Large models}} \\
 \hline
drama-large                                  & 400M   &          90 & 45.15            & 41.61          & 74.41                & 33.22          & 67.05          & 46.28          & 52.29               \\
 mmlw-roberta-large                           & 435M   &          96 & 66.15            & 44.58          & \cellcolor{good}{89.15}       & 49.91          & 85.23          & 61.58          & 67.00               \\
 mmlw-retrieval-roberta-large-v2              & 435M   &          80 & 64.62            & 39.08          & 86.53                & \cellcolor{good}{58.35} & \cellcolor{good}{85.64} & 63.09          & 66.84               \\
 mmlw-retrieval-roberta-large                 & 435M   &          93 & 63.90            & 45.18          & 88.48                & 57.23          & 84.71          & \cellcolor{good}{63.69} & \cellcolor{good}{67.90}       \\
 LaBSE                                        & 471M   &         100 & 57.35            & 42.40          & 79.27                & 27.36          & 74.67          & 48.52          & 56.21               \\
 KaLM-embedding-multilingual-mini-instruct-v1 & 494M   &          63 & 64.89            & 53.63          & 80.68                & 44.59          & 76.24          & 58.81          & 64.01               \\
 mmlw-e5-large                                & 560M   &          90 & 53.59            & 38.93          & 59.80                & 56.53          & 39.95          & 51.69          & 49.76               \\
 multilingual-e5-large                        & 560M   &          90 & 58.53            & 40.60          & 84.57                & 52.43          & 81.41          & 59.06          & 63.51               \\
 snowflake-arctic-embed-l-v2.0                & 568M   &          93 & 57.12            & 43.56          & 80.20                & 54.29          & 77.95          & 58.98          & 62.62               \\
 Qwen3-Embedding-0.6B                         & 596M   &          90 & \cellcolor{good}{69.66}   & \cellcolor{good}{56.65} & 81.31                & 48.59          & 78.45          & 62.20          & 66.93               \\
 \hline
 \multicolumn{8}{l}{\textbf{Extra large models (>1B)}} \\
 \hline
 drama-1b                                     & 1.2B  &          90 & 58.46            & 45.11             & 80.60                & 51.49             & 78.21             & 58.61             & 62.77               \\
 stella-pl                                    & 1.5B  &          80 & 66.94            & 38.08             & 89.20                & \cellcolor{second_best} \underline{60.82}             & \cellcolor{best} \textbf{86.87}    & 64.85             & 68.38               \\
 stella-pl-retrieval-8k                       & 1.5B  &          80 & 68.14            & 35.42             & \cellcolor{second_best} \underline{89.56}    & \cellcolor{best} \textbf{61.59}    & 86.56             & 64.98             & 68.25               \\
 Qwen3-Embedding-4B                           & 4.0B  &          90 & \cellcolor{second_best} \underline{79.30} & \cellcolor{best} \textbf{59.90}     & 86.68                & 56.65             & 85.55             & 69.37             & 73.62               \\
 Qwen3-Embedding-8B                           & 7.6B  &          90 & \cellcolor{best} \textbf{79.87}   & \cellcolor{second_best} \underline{58.64} & 87.61                & 59.21             & \cellcolor{second_best} \underline{86.72} & \cellcolor{best} \textbf{70.47}    & \cellcolor{best} \textbf{74.41}      \\
 BGE-Multilingual-Gemma2                      & 9.2B  &          83 & 77.77            & 58.15             & \cellcolor{best} \textbf{89.75}       & 58.93             & 83.97             &  \cellcolor{second_best} \underline{69.81} & \cellcolor{second_best} \underline{73.71}   \\
\hline
\end{tabular}
\caption{\label{tab:main-results}
Average of the main metric per task type and overall scores on PL-MTEB. The zero-shot column shows what percentage of the benchmark can be considered out-of-distribution for a given model. The best scores when considering models from the same size group are \colorbox{good}{highlighted}, the best scores among all models are marked in \colorbox{best}{\textbf{bold}}, and the second best are \colorbox{second_best}{\underline{underlined}}. 
}
\end{table*}

\section{Evaluation}
\label{sec:evaluation}

\subsection{Experimental setup}

The evaluation was conducted for the selected models using custom software\footnote{\href{https://github.com/rafalposwiata/pl-mteb}{https://github.com/rafalposwiata/pl-mteb}} built on the MTEB framework\footnote{\href{https://github.com/embeddings-benchmark/mteb}{https://github.com/embeddings-benchmark/mteb}}. Each model was run in accordance with the specifications provided by its developers or using a pre-existing implementation in MTEB. For each model, information about the datasets used for training was compiled where available. For models created using knowledge distillation, we have specified the datasets used to train the teacher model as the training datasets\footnote{For the clarity of the text, the teacher model’s training datasets will simply be referred to as “training datasets” and will not be distinguished from the actual datasets on which the models were trained.}. Information about the training datasets was used to determine the percentage of tasks in PL-MTEB that are new to the model—that is, the model was not trained on these or similar data, such as the English equivalents of the tasks we used. This has been added to the results tables as the zero-shot column, and had already been proposed in recent versions of the MTEB framework.
In the following subsections, we provide brief descriptions of the evaluated models, and then present and discuss the results.

\subsection{Models}

We run evaluations on dense embedding models trained in a supervised manner and that were recently state-of-the-art solutions. Below is a brief description of the evaluated models.

\noindent
\textbf{LaBSE} \citep{labse}\quad A language-agnostic BERT sentence embedding model supporting 109 languages optimized for bi-text mining tasks.

\noindent
\textbf{Multilingual SBERT} \citep{reimers-gurevych-2019-sentence}\quad Sentence-BERT (SBERT) is a modification of the pretrained BERT \citep{devlin-etal-2019-bert} network that use siamese and triplet network structures to generate text embeddings. In our experiments we used four multilingual SBERT models: distiluse-base-multilingual-cased-v2, paraphrase-multilingual-MiniLM-L12-v2, paraphrase-multilingual-mpnet-base-v2, and static-similarity-mrl-multilingual-v1.

\noindent
\textbf{Multilingual E5} \citep{wang2022text}\quad Text encoder supporting over 100 languages, developed using two-stage training procedure. The first stage involved weakly-supervised training on a dataset of text pairs extracted from large internet corpora, such as Common Crawl. In the second stage, the model was fine-tuned in a supervised manner on several annotated datasets. We used three versions of this model: small, base, and large.

\noindent
\textbf{KaLM-Embedding} \citep{kalm} A series of embedding models adapted from LLMs with superior training data. The KaLM-embedding-multilingual-mini-instruct-v1 model was trained from Qwen2-0.5B \citep{Yang2024Qwen2TR} using a two-stage approach similar to E5 models: massive weakly supervised pre-training and supervised fine-tuning.

\noindent
\textbf{Arctic-Embed 2.0} \citep{yu2024arcticembed20multilingualretrieval} Multilingual embedding models, trained using a multi-stage process similar to that described for the models mentioned earlier. For evaluation, we selected the snowflake-arctic-embed-m-v2.0 and snowflake-arctic-embed-l-v2.0 models, which are based on the gte-multilingual-base \citep{zhang2024mgte} and bge-m3-retromae \citep{bge-m3} models, respectively.

\noindent
\textbf{DRAMA} \citep{ma-etal-2025-drama} Dense retrieval models built upon a pruned LLM backbone and fine-tuned on diverse LLM-augmented data in a single-stage contrastive learning setup. We evaluated three versions of this model: base, large, and 1b.

\noindent
\textbf{Qwen3-Embedding} \citep{qwen3embedding} A model series specifically designed for text embedding and ranking tasks. Models are based on Qwen3 \citep{Yang2025Qwen3TR} and trained using a multistage pipeline that combines large-scale weakly supervised pre-training, supervised fine-tuning on high-quality synthetic data, and checkpoints merging.  We evaluated models in three sizes: 0.6B, 4B, and 8B.

\noindent
\textbf{BGE-Multilingual-Gemma2} \citep{c-mteb, bge-m3} A multilingual embedding model based on Gemma-2-9b \citep{gemmateam2024}. It was trained on a diverse range of tasks such as retrieval, classification, and clustering in various languages. 

\noindent
\textbf{Silver Retriever} \citep{rybak-ogrodniczuk-2024-silver-retriever}\quad Polish dense retrieval model trained on MAUPQA \citep{rybak2023maupqa} - manually or weakly labeled datasets.. The model was based on the HerBERT language model \citep{mroczkowski2021herbert}.

\noindent
\textbf{Polish SBERT} \citep{dadas-para}\quad SBERT model trained using multilingual knowledge distillation technique \citep{reimers-gurevych-2020-making} and Polish-English bilingual corpus. In our experiments we used two such models:  st-polish-paraphrase-from-mpnet and st-polish-paraphrase-from-distilroberta.

\noindent
\textbf{MMLW} \citep{dadas-etal-2024-pirb-comprehensive}\quad A set of models trained using a bilingual Polish-English  corpus and the knowledge distillation technique. The authors selected two groups of models as student models: pre-trained Polish RoBERTa language models \citep{dadas2020pre} and multilingual E5 \citep{wang2022text}. As teachers, they chose English BGE \citep{c-mteb} models. For experiments, we used five models prepared in that way:  mmlw-roberta-base, mmlw-roberta-large, mmlw-e5-small,
mmlw-e5-base, and mmlw-e5-large. In addition, we tested two mmlw models designed for retrieval: mmlw-retrieval-roberta-large and mmlw-retrieval-roberta-large-v2. Version 2 of the model was trained using a different teacher model, namely, stella\_en\_1.5B\_v5 \citep{zhang2025jasperstelladistillationsota}, and fine-tuned on a larger dataset of over 4 million queries, whereas first version, which used only the Polish MSMSRCO \citep{wojtasik2023beirpl} dataset.

\noindent
\textbf{Stella-PL} \citep{dadas-etal-2024-pirb-comprehensive} Bilingual Polish-English text encoders based on stella\_en\_1.5B\_v5 adapted for Polish with a multilingual knowledge distillation method using a diverse corpus of 20 million Polish-English text pairs. 
Model stella-pl-retrieval-8k has an extended context and was fine-tuned for retrieval using a dataset comprising 1.5 million queries. 

\subsection{Main Results}
\label{sec:main_results}
The main results of our experiments are presented in Table \ref{tab:main-results}. The \textbf{Qwen3-Embedding-8B} model achieved the best overall score for the entire benchmark as measured by the average across all tasks as well as by task type. Its advantage over other models, particularly Polish ones, was mainly due to its strong performance on classification and clustering tasks. At the same time, its results for the other task types did not differ significantly from the best ones. Considering the results by task type, none of the models performed best for more than one type. Generally, the largest models with over 1 billion parameters achieved the best results, as expected.
However, it should be noted that most of the models we evaluated, including those with the best performance, had data in their training sets that were, to some extent, similar to the data in our benchmark, as shown in the zero-shot column.
In the following subsections, we will analyze the results for each task type and then identify which models perform best in each size group.

\subsection{Results by Task Type}
\label{sec:results_by_task_type}
\paragraph{Classification} The best results were achieved by models from the Qwen3-Embeddings family, specifically \textbf{Qwen3-Embedding-8B} and \textbf{Qwen3-Embedding-4B}. Looking at the detailed results in Table \ref{tab:classification-results}, the \textbf{Qwen3-Embedding-8B} model performed best on five tasks. An interesting case is the PAC task, where the best result was achieved by the very compact \textbf{multilingual-e5-small} model. Apart from the \textbf{KaLM-embedding-multilingual-mini-instruct-v1} model, the other models had a zero-shot score of 100, meaning they did not use similar classification tasks for training. 

\paragraph{Clustering} In the clustering tasks, the same models that performed best in classification, namely \textbf{Qwen3-Embedding-8B} and \textbf{Qwen3-Embedding-4B}, achieved the best results. According to the detailed results in Table \ref{tab:custering-results}, this time the smaller \textbf{Qwen3-Embedding-4B} model performed better, winning the EightTags task and two variants of the WikinewsPL task. In hierarchical clustering, the \textbf{BGE-Multilingual-Gemma2} model performed best. Comparing the results for the tasks we proposed, the models perform better on P2P tasks than on S2S tasks. P2P tasks contain longer texts and, consequently, a greater amount of information used for proper grouping. Furthermore, creating tasks on new datasets ensured that none of the models used similar data during training, as illustrated by the zero-shot column.

\paragraph{Pair Classification} In this type of task, the \textbf{BGE-Multilingual-Gemma2} model achieved the best average score. Analyzing the detailed results in Table \ref{tab:pairclassification-results}, this model performed best on two tasks. In the remaining two tasks, very good results were achieved by Polish models, among which the \textbf{stella-pl-retrieval-8k} model stands out as one of the three winners in the PSC task, and second in the pair classification category. As in clustering, all models had a zero-shot score of 100.

\paragraph{Retrieval} In this largest group of tasks, the best average score was achieved by the \textbf{stella-pl-retrieval-8k} model, followed closely by the \textbf{stella-pl} model. The detailed results presented in Table \ref{tab:retrieval-results} show that these models either won or placed second in most retrieval tasks. It should be noted that this result was influenced by the fact that the training data for these models included similar tasks from this category. At the same time, most models used some of these tasks during training, such as the MSMARCO \citep{Campos2016MSMA} training split, which is common practice. 

\paragraph{Semantic Textual Similarity (STS)} For tasks of this type, the \textbf{stella-pl} model achieved the highest average score, slightly outperforming the \textbf{Qwen3-Embedding-8B} model. The results for specific tasks shown in Table \ref{tab:sts-results} indicate that each of these models won only one task. It should be noted that during the training of the stella-pl teacher model, the STSBenchmark task, which is the English version of the STSBenchmarkMultilingual task, was used. However, the model did not achieve the best result for this task.

\subsection{Results by Model Size}

From a practical standpoint, when resources are often limited, the goal is to find a solution that is both scalable and delivers good results. In the following paragraphs, we analyze the results of our benchmark, taking model size into consideration.

\paragraph{Small models (< 150M)} 

Among the smallest models, the \textbf{mmlw-roberta-base}\footnote{The name of this model suggests that it belongs to the “base” group, but with 124M parameters, it is actually better suited to the “small” group.} model achieved significantly better results than other models, both in average scores across the entire benchmark and on individual task types, ranking second only in clustering and winning in all other categories.

\paragraph{Base models} 

There is no clear winner in this group of models. We can highlight the \textbf{snowflake-arctic-embed-m-v2.0} model, which achieved the best average score across all tasks without being the best in any single task type, and the \textbf{multilingual-e5-base} model, which achieved the best average scores by type and was the best in classification and clustering. At the same time, these models performed mostly worse than the best model from the previous group, namely \textbf{mmlw-roberta-base}.

\paragraph{Large models} 

In this group as well, no single model has a clear advantage over the others. The \textbf{mmlw-retrieval-roberta-large} model achieved the best average results, though it did not outperform the others on any specific task type. Looking at individual task types, the second version of this model, \textbf{mmlw-retrieval-roberta-large-v2}, achieved the best results for retrieval and STS tasks. However, it should be noted that for these types, the zero-shot scores for this model are 54 and 66, respectively, indicating the use of similar data during training. For classification and clustering tasks, the best results were achieved by \textbf{Qwen3-Embedding-0.6B}, the smallest of the tested models from the Qwen3-Embedding family. 

\paragraph{Extra large models (> 1B)} 

As described in subsections \ref{sec:main_results} and \ref{sec:results_by_task_type}, the best results in our benchmark were achieved by very large models with over 1 billion parameters. Although the \textbf{Qwen3-Embedding-8B} and \textbf{BGE-Multilingual-Gemma2}  models achieved the best average results, they outperformed others in only a single category across the various task types. 

\section{Conclusion}
\label{sec:conclusion}
In this work, we introduce PL-MTEB, a text embedding benchmark for the Polish language comprising 30 tasks across 5 categories.
We evaluated 30 models, including Polish and multilingual ones. The \textbf{Qwen3-Embedding-8B} achieved the best average result. The results indicate that there is no single universal model that performs best across all task types. Considering model size, among the smallest models, the \textbf{mmlw-roberta-base} model achieved very good results, outperforming larger models from the next size group. On the other hand, the results were influenced by the fact that some models used similar data during training, particularly in retrieval tasks. We believe that our work will help standardize the evaluation of text embedding models for Polish. At the same time, tasks from PL-MTEB can be used by the broader international community to improve the accuracy of evaluations of multilingual embeddings. PL-MTEB is a benchmark that will be successively updated with results for new models. Given the public nature of our benchmark and the findings related to zero-shot settings, we plan to expand the benchmark to include closed tasks in the future.

The source code for evaluating new models or reproducing our experiments is available at \url{https://github.com/rafalposwiata/pl-mteb}. Datasets and public leaderboard can be found at \url{https://huggingface.co/PL-MTEB}. As PL-MTEB is part of the MTEB project, the source code of the tasks themselves and details related to the evaluation are at \url{https://github.com/embeddings-benchmark/mteb}. 

\section*{Limitations}

\paragraph{Long document datasets} The tasks in PL-MTEB are based on texts of varying lengths, but most are short or medium-length. There are no tasks involving very long texts, which is often the case in real-world applications, such as RAG systems.

\paragraph{Limited conclusions for specific domains} The selection of tasks limits the conclusions that can be drawn about the model’s performance in specialized fields such as law or finance, as PL-MTEB contains only one dataset for each field. It would be useful to expand the benchmark to include tasks from these fields.

\paragraph{Closed-source models} We evaluated only publicly available models, excluding closed ones accessible via API, such as text-embedding-3-small from OpenAI. This was due to the limited budget of the project. We plan to include such solutions in the future.

\section*{Acknowledgments}

We want to thank all contributors to the MTEB project, whose work and support enabled us to create our benchmark, and in particular, the project leaders Niklas Muennighoff and Kenneth Enevoldsen.



\bibliography{references}

\appendix

\section{Tasks Descriptions}
\label{sec:tasks_appendix}

\subsection{Classification}

\noindent
\textbf{CBD} \citep{ptaszynski2019results}\quad The Cyberbullying Detection task, where the goal is to predict if tweet contains a cyberbullying content.

\noindent
\textbf{PAC} \citep{NEURIPS2022_890b206e}\quad Polish Abusive Clauses Dataset used to formulate binary classification task of detecting abusive clauses.

\noindent
\textbf{PolEmo2.0-IN} and \textbf{PolEmo2.0-OUT} \citep{kocon-etal-2019-multi}\quad Based on a collection of Polish online reviews from four domains: medicine, hotels, products and school. The PolEmo2.0-IN task is to predict the sentiment of in-domain (medicine and hotels) reviews. The PolEmo2.0-OUT task is to predict the sentiment of out-of-domain (products and school) reviews using models train on reviews from medicine and hotels domains.

\noindent
\textbf{MassiveIntent} and \textbf{MassiveScenario} \citep{fitzgerald2022massive}\quad The tasks include intent and scenario detection from the content of utterances addressed to Amazon's Alexa virtual assistant. They are based on a multilingual dataset with 51 available languages, of which we used only Polish-language subset. The tasks were already in MTEB.

\noindent
\textbf{AllegroReviews} \citep{rybak-etal-2020-klej}\quad Based on a Polish dataset for sentiment classification on reviews from e-commerce marketplace Allegro. The task is to predict a rating ranging from 1 to 5.

\subsection{Clustering}

\noindent
\textbf{EightTags} (original name 8Tags) \citep{dadas-etal-2020-evaluation}\quad Clustering of headlines from social media posts in Polish belonging to 8 categories: film, history, food, medicine, motorization, work, sport and technology.

\noindent
\textbf{PlscHierarchicalS2S} and \textbf{PlscHierarchicalP2P}\quad Tasks involve clustering publication titles and titles with abstracts, respectively, first in terms of their scientific field and than by scientific disciplines.

\noindent
\textbf{WikinewsPLS2S} and \textbf{WikinewsPLP2P}\quad Tasks involve clustering Wikinews article titles and titles with texts, respectively, in terms of category.

\subsection{Pair Classification}

\noindent
\textbf{SICK-E-PL} \citep{dadas-etal-2020-evaluation}\quad 
The binary variant of textual entailment task based on the Polish version of Sentences Involving Compositional Knowledge (SICK) \citep{marelli-etal-2014-sick} dataset, where labels 'neutral' and 'contradiction' was merged to create one 'not entailed' class.

\noindent
\textbf{CDSC-E} \citep{wroblewska2017polish}\quad 
The binary variant of textual entailment task based on Compositional Distributional Semantics Corpus, where labels 'neutral' and 'contradiction' was merged to create one 'not entailed' class.

\noindent
\textbf{PPC} \citep{dadas-para}\quad A task to detect whether a given sentence is a paraphrase of another. Based on a Polish Paraphrase Corpus, class 'exact paraphrase' and 'close paraphrase' are merged.

\noindent
\textbf{PSC} \citep{ogro:kop:14:lrec}\quad The task is to detect whether two summaries relate to the same article. Base on The Polish Summaries Corpus.

\begin{table*}[h]
    \small
    \centering
    \begin{tabular}{l|l}
    \hline
    \bf Name in Paper &  \bf HF Name \\
    \hline
   LaBSE & sentence-transformers/LaBSE \\
distiluse-base-multilingual-cased-v2 & sentence-transformers/distiluse-base-multilingual-cased-v2 \\
paraphrase-multilingual-MiniLM-L12-v2 & sentence-transformers/paraphrase-multilingual-MiniLM-L12-v2 \\
paraphrase-multilingual-mpnet-base-v2 & sentence-transformers/paraphrase-multilingual-mpnet-base-v2 \\
static-similarity-mrl-multilingual-v1 & sentence-transformers/static-similarity-mrl-multilingual-v1 \\
multilingual-e5-small & intfloat/multilingual-e5-small \\
multilingual-e5-base & intfloat/multilingual-e5-base \\
multilingual-e5-large & intfloat/multilingual-e5-large \\
KaLM-embedding-multilingual-mini-instruct-v1 & HIT-TMG/KaLM-embedding-multilingual-mini-instruct-v1 \\
snowflake-arctic-embed-l-v2.0 & Snowflake/snowflake-arctic-embed-l-v2.0 \\
snowflake-arctic-embed-m-v2.0 & Snowflake/snowflake-arctic-embed-m-v2.0 \\
drama-base & facebook/drama-base \\
drama-large & facebook/drama-large \\
drama-1b & facebook/drama-1b \\
Qwen3-Embedding-0.6B & Qwen/Qwen3-Embedding-0.6B \\
Qwen3-Embedding-4B & Qwen/Qwen3-Embedding-4B \\
Qwen3-Embedding-8B & Qwen/Qwen3-Embedding-8B \\
BGE-Multilingual-Gemma2 & BAAI/bge-multilingual-gemma2 \\
silver-retriever-base-v1.1 & ipipan/silver-retriever-base-v1.1 \\
st-polish-paraphrase-from-mpnet & sdadas/st-polish-paraphrase-from-mpnet \\
st-polish-paraphrase-from-distilroberta & sdadas/st-polish-paraphrase-from-distilroberta \\
mmlw-e5-small & sdadas/mmlw-e5-small \\
mmlw-e5-base & sdadas/mmlw-e5-base \\
mmlw-e5-large & sdadas/mmlw-e5-large \\
mmlw-roberta-base & sdadas/mmlw-roberta-base \\
mmlw-roberta-large & sdadas/mmlw-roberta-large \\
mmlw-retrieval-roberta-large & sdadas/mmlw-retrieval-roberta-large \\
mmlw-retrieval-roberta-large-v2 & sdadas/mmlw-retrieval-roberta-large-v2 \\
stella-pl & sdadas/stella-pl \\
stella-pl-retrieval-8k & sdadas/stella-pl-retrieval-8k \\
    \hline
    \end{tabular}
    \caption{Model names as referenced in the paper, and corresponding Hugging Face Hub identifiers.}
    \label{tab:model_links}
\end{table*}

\subsection{Retrieval}

The vast majority of retrieval tasks are from BEIR-PL \citep{wojtasik2023beirpl}, which was created by automatic translating dataset from BEIR \citep{thakur2021beir} to Polish language.

\noindent
\textbf{ArguAna-PL}\quad Retrieving the best counterargument to a given argument.

\noindent
\textbf{DBPedia-PL}\quad Searching for entities in the DBpedia knowledge base.

\noindent
\textbf{FiQA-PL}\quad Retrieving relevant documents from financial domain to a given query.

\noindent
\textbf{HotpotQA-PL}\quad A question answering task which requires reasoning over multiple paragraphs (multi-hop) and Wikipedia articles are the information source.

\noindent
\textbf{MSMARCO-PL}\quad A question answering task based on Bing questions and human generated answers.

\noindent
\textbf{NFCorpus-PL}\quad Retrieving relevant documents from NutrionFacts (medicine domain) to a given query.

\noindent
\textbf{NQ-PL}\quad A question answering task where the questions are from a Google search engine and the answers are annotated by a human based on Wikipedia articles.

\noindent
\textbf{Quora-PL}\quad Task is based on questions that are marked as duplicates on the Quora platform. Given a question, find other (duplicate) questions.

\noindent
\textbf{SCIDOCS-PL}\quad Citation prediction task, where the goal is to get cited scientific articles based on the title of the article that cites them.

\noindent
\textbf{SciFact-PL}\quad Verifing scientific claims using evidence from the research literature containing scientific paper abstracts.

\noindent
\textbf{TRECCOVID-PL}\quad Retrieving relevant scientific articles related to COVID-19 based on a given query.

\subsection{Semantic Textual Similarity (STS)}

\noindent
\textbf{SICK-R-PL} \citep{dadas-etal-2020-evaluation}\quad  Textual relatedness task based on Polish version of Sentences Involving Compositional Knowledge (SICK) \citep{marelli-etal-2014-sick} dataset. 

\noindent
\textbf{CDSC-R} \citep{wroblewska2017polish}\quad Textual relatedness task based on Compositional Distributional Semantics Corpus.

\noindent
\textbf{STSBenchmarkMultilingual} Semantic Textual Similarity Benchmark (STSbenchmark) dataset, translated using DeepL API. Source of the dataset: \url{https://github.com/PhilipMay/stsb-multi-mt}. We used only Polish-language subset. The task was already in MTEB.

\section{Models}

Table \ref{tab:model_links} contains references to the evaluated models.

\section{Results}

Detailed results for each type of task are presented in Tables \ref{tab:classification-results}--\ref{tab:sts-results}. These results show, among other things, that most models were trained on retrieval data, which is why the zero-shot score for these models is less than 100\%.

\begin{table*}[h]
    \scriptsize
    \centering
    \begin{tabular}{lr|llllllll}
\hline
 \bf Model name  &  \rot{\bf Zero shot} & \rot{\bf CBD} &   \rot{\bf PolEmo2.0-IN} &    \rot{\bf PolEmo2.0-OUT} &    \rot{\bf AllegroReviews} &    \rot{\bf PAC} &    \rot{\bf MassiveIntent} &    \rot{\bf MassiveScenario} & \bf Avg. \\
\hline
 \multicolumn{8}{l}{\textbf{Small models (< 150M)}} \\
 \hline
static-similarity-mrl-multilingual-v1        &         100 & 54.19            & 53.38             & 38.40              & 26.40               & 56.63             & 53.78                         & 54.40                           & 48.17            \\
 paraphrase-multilingual-MiniLM-L12-v2        &         100 & 53.56            & 59.24             & 28.34              & 31.10               & 62.77             & 59.54                         & 65.16                           & 51.39            \\
 multilingual-e5-small                        &         100 & 58.22            & 58.05             & 24.28              & 35.35               & \cellcolor{best} \textbf{71.03}    & 57.96                         & 63.58                           & 52.64            \\
 mmlw-e5-small                                &         100 & 60.87            & 70.11             & 47.24              & 35.23               & 64.82             & 69.66                         & 72.90                           & 60.12            \\
 silver-retriever-base-v1.1                   &         100 & 63.36            & 62.60             & 43.31              & 33.57               & 61.68             & 66.45                         & 68.27                           & 57.03            \\
 st-polish-paraphrase-from-mpnet              &         100 & 67.30            & 67.83             & 31.62              & 35.35               & 63.13             & 66.04                         & 71.75                           & 57.57            \\
 st-polish-paraphrase-from-distilroberta      &         100 & 64.96            & 66.02             & 40.97              & 33.27               & 63.46             & 65.09                         & 70.17                           & 57.71            \\
 mmlw-roberta-base                            &         100 & 63.15            & 73.03             & 47.81              & 39.82               & 65.86             & 72.55                         & 75.50                           & 62.53            \\
 distiluse-base-multilingual-cased-v2         &         100 & 51.94            & 51.09             & 32.29              & 28.69               & 64.63             & 52.85                         & 61.15                           & 48.95            \\
  \hline
 \multicolumn{8}{l}{\textbf{Base models}} \\
 \hline
 drama-base                                   &         100 & 49.38            & 52.59             & 23.59              & 28.12               & 58.41             & 37.31                         & 44.99                           & 42.06            \\
 mmlw-e5-base                                 &         100 & 52.93            & 47.40             & 34.50              & 25.13               & 62.82             & 53.09                         & 55.74                           & 47.37            \\
 paraphrase-multilingual-mpnet-base-v2        &         100 & 57.77            & 62.78             & 19.76              & 36.19               & 62.48             & 64.75                         & 68.87                           & 53.23            \\
 multilingual-e5-base                         &         100 & 57.35            & 58.88             & 35.80              & 37.76               & 70.09             & 61.82                         & 65.79                           & 55.36            \\
 snowflake-arctic-embed-m-v2.0                &         100 & 62.52            & 58.20             & 28.17              & 29.89               & 64.97             & 64.84                         & 69.51                           & 54.01            \\
 \hline
 \multicolumn{8}{l}{\textbf{Large models}} \\
 \hline
 drama-large                                  &         100 & 53.61            & 52.99             & 24.14              & 28.73               & 60.23             & 44.22                         & 52.12                           & 45.15            \\
 mmlw-retrieval-roberta-large                 &         100 & 65.13            & 70.50             & 52.68              & 41.00               & 63.67             & 76.14                         & 78.17                           & 63.90            \\
 mmlw-retrieval-roberta-large-v2              &         100 & 62.89            & 75.98             & 55.15              & 40.92               & 67.84             & 72.50                         & 77.07                           & 64.62            \\
 mmlw-roberta-large                           &         100 & 64.44            & 77.58             & 55.60              & 47.24               & 65.33             & 75.13                         & 77.74                           & 66.15            \\
 LaBSE                                        &         100 & 64.69            & 64.56             & 47.24              & 35.44               & 65.58             & 59.83                         & 64.12                           & 57.35            \\
 KaLM-embedding-multilingual-mini-instruct-v1 &          71 & 61.35            & 78.61             & 61.36              & 56.30               & 62.13             & 62.49                         & 71.99                           & 64.89            \\
 mmlw-e5-large                                &         100 & 50.72            & 63.60             & 42.74              & 34.65               & 65.83             & 56.23                         & 61.36                           & 53.59            \\
 multilingual-e5-large                        &         100 & 61.50            & 65.58             & 38.17              & 39.21               & \cellcolor{second_best} \underline{70.48} & 66.07                         & 68.67                           & 58.53            \\
 snowflake-arctic-embed-l-v2.0                &         100 & 65.22            & 62.51             & 34.71              & 31.87               & 64.96             & 68.22                         & 72.38                           & 57.12            \\
 Qwen3-Embedding-0.6B                         &         100 & 63.42            & 87.42             & 71.74              & 59.88               & 61.60             & 70.38                         & 73.18                           & 69.66            \\
 \hline
 \multicolumn{8}{l}{\textbf{Extra large models (>1B)}} \\
 \hline
 drama-1b                                     &         100 & 59.45            & 68.31             & 47.38              & 40.62               & 63.43             & 62.72                         & 67.29                           & 58.46            \\
 stella-pl                                    &         100 & 65.19            & 82.05             & 60.28              & 48.07               & 62.35             & 73.50                         & 77.16                           & 66.94            \\
 stella-pl-retrieval-8k                       &         100 & 67.17            & 82.80             & 64.00              & 48.48               & 63.51             & 74.02                         & 77.00                           & 68.14            \\
 Qwen3-Embedding-4B                           &         100 & 81.41            & 90.37             & 77.73              & \cellcolor{second_best} \underline{68.95}   & 69.89             & \cellcolor{second_best} \underline{81.24}             & \cellcolor{second_best} \underline{85.5}                & \cellcolor{second_best} \underline{79.3} \\
 Qwen3-Embedding-8B                           &         100 & \cellcolor{best} \textbf{83.71}   & \cellcolor{second_best} \underline{91.29} & \cellcolor{best} \textbf{79.41}     & \cellcolor{best} \textbf{69.37}      & 65.22             & \cellcolor{best} \textbf{83.11}                & \cellcolor{best} \textbf{86.96}                  & \cellcolor{best} \textbf{79.87}   \\
 BGE-Multilingual-Gemma2                      &         100 & \cellcolor{second_best} \underline{82.6} & \cellcolor{best} \textbf{91.63}    & \cellcolor{second_best} \underline{78.4}   & 64.53               & 66.16             & 79.52                         & 81.57                           & 77.77            \\
\hline
\end{tabular}
\caption{\label{tab:classification-results}
Evaluation results on classification tasks using accuracy metric. The best scores for a given column are marked in \colorbox{best}{\textbf{bold}}, and the second best are \colorbox{second_best}{\underline{underlined}}. 
}
\end{table*}

\begin{table*}[h]
    \scriptsize
    \centering
    \begin{tabular}{lr|llllllll}
    \hline
\bf Model name  &   \rot{\bf Zero shot} & \rot{\bf EightTags}   & \rot{\bf PlscHierarchicalS2S}   & \rot{\bf PlscHierarchicalP2P}   & \rot{\bf WikinewsPLS2S}   &  \rot{\bf WikinewsPLP2P}   &  \bf Avg.           \\
\hline
 \multicolumn{7}{l}{\textbf{Small models (< 150M)}} \\
 \hline
 static-similarity-mrl-multilingual-v1        &         100 & 16.93                    & 37.57                           & 46.63                           & 19.01                     & 30.08                     & 30.04             \\
 paraphrase-multilingual-MiniLM-L12-v2        &         100 & 26.14                    & 47.64                           & 54.75                           & 30.54                     & 44.33                     & 40.68             \\
 multilingual-e5-small                        &         100 & 30.21                    & 49.83                           & 55.88                           & 41.48                     & 42.55                     & 43.99             \\
 mmlw-e5-small                                &         100 & 32.28                    & 52.40                           & 56.96                           & 44.66                     & 58.25                     & 48.91             \\
 st-polish-paraphrase-from-distilroberta      &         100 & 30.40                    & 47.47                           & 55.78                           & 35.30                     & 44.58                     & 42.71             \\
 st-polish-paraphrase-from-mpnet              &         100 & 31.30                    & 49.31                           & 56.94                           & 37.65                     & 47.43                     & 44.53             \\
 silver-retriever-base-v1.1                   &         100 & 32.18                    & 49.19                           & 56.88                           & 39.68                     & 46.67                     & 44.92             \\
 mmlw-roberta-base                            &         100 & 31.61                    & 51.02                           & 58.35                           & 46.11                     & 52.89                     & 48.00             \\
 distiluse-base-multilingual-cased-v2         &         100 & 26.90                    & 41.98                           & 51.42                           & 31.78                     & 42.20                     & 38.86             \\
  \hline
 \multicolumn{8}{l}{\textbf{Base models}} \\
 \hline
 drama-base                                   &         100 & 24.90                    & 47.28                           & 53.22                           & 28.22                     & 48.79                     & 40.48             \\
 mmlw-e5-base                                 &         100 & 23.72                    & 44.23                           & 53.88                           & 26.47                     & 38.14                     & 37.29             \\
 paraphrase-multilingual-mpnet-base-v2        &         100 & 29.41                    & 48.89                           & 51.52                           & 32.81                     & 44.08                     & 41.34             \\
 multilingual-e5-base                         &         100 & 31.17                    & 49.67                           & 53.63                           & 40.94                     & 45.11                     & 44.10             \\
 snowflake-arctic-embed-m-v2.0                &         100 & 30.12                    & 49.94                           & 54.42                           & 41.75                     & 42.77                     & 43.80             \\
 \hline
 \multicolumn{8}{l}{\textbf{Large models}} \\
 \hline
 drama-large                                  &         100 & 26.98                    & 48.98                           & 53.48                           & 29.40                     & 49.21                     & 41.61             \\
 mmlw-retrieval-roberta-large-v2              &         100 & 27.53                    & 47.49                           & 51.97                           & 32.33                     & 36.07                     & 39.08             \\
 mmlw-roberta-large                           &         100 & 33.35                    & 53.66                           & 56.97                           & 34.93                     & 43.98                     & 44.58             \\
 mmlw-retrieval-roberta-large                 &         100 & 31.79                    & 51.66                           & 55.47                           & 41.22                     & 45.74                     & 45.18             \\
 LaBSE                                        &         100 & 26.11                    & 48.45                           & 57.06                           & 35.40                     & 44.99                     & 42.40             \\
 KaLM-embedding-multilingual-mini-instruct-v1 &         100 & 38.84                    & 52.63                           & 60.89                           & 55.67                     & \cellcolor{second_best} \underline{60.14}         & 53.63             \\
 mmlw-e5-large                                &         100 & 27.93                    & 45.04                           & 55.39                           & 27.30                     & 39.01                     & 38.93             \\
 multilingual-e5-large                        &         100 & 27.18                    & 50.49                           & 53.74                           & 31.13                     & 40.46                     & 40.60             \\
 snowflake-arctic-embed-l-v2.0                &         100 & 33.47                    & 51.64                           & 55.52                           & 38.00                     & 39.17                     & 43.56             \\
 Qwen3-Embedding-0.6B                         &         100 & 46.65                    & 55.47                           & \cellcolor{second_best} \underline{62.56}               & \cellcolor{second_best} \underline{59.21}         & 59.36                     & 56.65             \\
 \hline
 \multicolumn{8}{l}{\textbf{Extra large models (>1B)}} \\
 \hline
 drama-1b                                     &         100 & 33.18                    & 51.56                           & 54.76                           & 37.49                     & 48.54                     & 45.11             \\
 stella-pl-retrieval-8k                       &         100 & 23.23                    & 43.30                           & 48.40                           & 28.17                     & 34.00                     & 35.42             \\
 stella-pl                                    &         100 & 23.20                    & 45.82                           & 52.34                           & 27.58                     & 41.45                     & 38.08             \\
 Qwen3-Embedding-4B                           &         100 & \cellcolor{best} \textbf{62.30}            & \cellcolor{second_best} \underline{56.57}               & 60.69                           & \cellcolor{best} \textbf{59.62}            & \cellcolor{best} \textbf{60.30}             & \cellcolor{best} \textbf{59.90}     \\
 Qwen3-Embedding-8B                           &         100 & \cellcolor{second_best} \underline{60.4}         & 56.19                           & 61.22                           & 55.74                     & 59.63                     & \cellcolor{second_best} \underline{58.64} \\
 BGE-Multilingual-Gemma2                      &         100 & 59.27                    & \cellcolor{best} \textbf{58.68}                  & \cellcolor{best} \textbf{62.95}                  & 54.01                     & 55.82                     & 58.15             \\
\hline
\end{tabular}
\caption{\label{tab:custering-results}
Evaluation results on clustering tasks using v-measure. The best scores for a given column are marked in \colorbox{best}{\textbf{bold}}, and the second best are \colorbox{second_best}{\underline{underlined}}. 
}
\end{table*}

\begin{table*}[h]
    \scriptsize
    \centering
    \begin{tabular}{lr|lllll}
\hline
\bf Model name  &   \rot{\bf Zero shot} & \rot{\bf SICK-E-PL}     & \rot{\bf CDSC-E}         &  \rot{\bf PSC}       &  \rot{\bf PPC}            & \bf Avg.           \\
\hline
    \multicolumn{7}{l}{\textbf{Small models (< 150M)}} \\
 \hline
 static-similarity-mrl-multilingual-v1        &         100 & 53.92             & 57.82             & 95.33          & 74.57             & 70.41             \\
 multilingual-e5-small                        &         100 & 67.48             & 72.18             & 99.40          & 87.74             & 81.70             \\
 paraphrase-multilingual-MiniLM-L12-v2        &         100 & 71.78             & 72.39             & 97.07          & 92.37             & 83.40             \\
 mmlw-e5-small                                &         100 & 77.49             & \cellcolor{second_best} \underline{79.34} & 98.17          & 91.68             & 86.67             \\
 silver-retriever-base-v1.1                   &         100 & 55.84             & 62.67             & 98.75          & 82.04             & 74.82             \\
 st-polish-paraphrase-from-distilroberta      &         100 & 79.41             & 76.03             & 99.09          & 93.31             & 86.96             \\
 st-polish-paraphrase-from-mpnet              &         100 & 80.39             & 75.17             & 99.03          & 93.67             & 87.06             \\
 mmlw-roberta-base                            &         100 & 81.85             & 79.23             & 98.59          & 92.97             & 88.16             \\
 distiluse-base-multilingual-cased-v2         &         100 & 62.29             & 72.10             & 96.26          & 86.83             & 79.37             \\
  \hline
 \multicolumn{7}{l}{\textbf{Base models}} \\
 \hline
 drama-base                                   &         100 & 54.05             & 60.18             & 95.53          & 78.45             & 72.05             \\
 mmlw-e5-base                                 &         100 & 42.69             & 43.76             & 78.91          & 72.64             & 59.50             \\
 multilingual-e5-base                         &         100 & 68.52             & 72.23             & 99.28          & 88.30             & 82.08             \\
 paraphrase-multilingual-mpnet-base-v2        &         100 & 77.07             & 75.88             & 98.22          & 93.67             & 86.21             \\
 snowflake-arctic-embed-m-v2.0                &         100 & 59.57             & 70.24             & \cellcolor{best} \textbf{99.54} & 84.13             & 78.37             \\
 \hline
 \multicolumn{7}{l}{\textbf{Large models}} \\
 \hline
 drama-large                                  &         100 & 57.47             & 64.14             & 95.77          & 80.26             & 74.41             \\
 mmlw-retrieval-roberta-large-v2              &         100 & 79.27             & 75.61             & \cellcolor{best} \textbf{99.54} & 91.69             & 86.53             \\
 mmlw-retrieval-roberta-large                 &         100 & 83.15             & 78.53             & 99.42          & 92.81             & 88.48             \\
 mmlw-roberta-large                           &         100 & 84.29             & \cellcolor{best} \textbf{79.96}    & 98.80          & 93.56             & 89.15             \\
 LaBSE                                        &         100 & 63.67             & 69.06             & 97.37          & 86.97             & 79.27             \\
 KaLM-embedding-multilingual-mini-instruct-v1 &         100 & 63.78             & 71.63             & 99.48          & 87.81             & 80.68             \\
 mmlw-e5-large                                &         100 & 43.30             & 37.10             & 80.53          & 78.26             & 59.80             \\
 multilingual-e5-large                        &         100 & 75.42             & 72.28             & 99.43          & 91.16             & 84.57             \\
 snowflake-arctic-embed-l-v2.0                &         100 & 63.24             & 71.02             & 99.48          & 87.08             & 80.20             \\
 Qwen3-Embedding-0.6B                         &         100 & 68.29             & 68.87             & 97.85          & 90.22             & 81.31             \\
 \hline
 \multicolumn{7}{l}{\textbf{Extra large models (>1B)}} \\
 \hline
 drama-1b                                     &         100 & 66.32             & 70.11             & 99.38          & 86.60             & 80.60             \\
 stella-pl                                    &         100 & 84.68             & 79.20             & 99.31          & 93.60             & 89.20             \\
 stella-pl-retrieval-8k                       &         100 & \cellcolor{second_best} \underline{85.66} & 79.26             & \cellcolor{best} \textbf{99.54} & 93.77             & \cellcolor{second_best} \underline{89.56} \\
 Qwen3-Embedding-4B                           &         100 & 79.82             & 73.59             & 98.68          & 94.61             & 86.68             \\
 Qwen3-Embedding-8B                           &         100 & 82.47             & 74.84             & 98.43          & \cellcolor{second_best} \underline{94.71} & 87.61             \\
 BGE-Multilingual-Gemma2                      &         100 & \cellcolor{best} \textbf{85.8}     & 78.51             & 99.27          & \cellcolor{best} \textbf{95.43}    & \cellcolor{best} \textbf{89.75}    \\
 \hline
 \end{tabular}
\caption{\label{tab:pairclassification-results}
Evaluation results on pair classification tasks using average precision score based on cosine similarity. The best scores for a given column are marked in \colorbox{best}{\textbf{bold}}, and the second best are \colorbox{second_best}{\underline{underlined}}. 
}
\end{table*}

\begin{table*}[h]
    \scriptsize
    \centering
\setlength{\tabcolsep}{0.6em}
\begin{tabular}{lr|llllllllllll}
\hline
 \bf Model name  &   \rot{\bf Zero shot} & \rot{\bf ArguAna-PL}        & \rot{\bf DBPedia-PLHardNeg}  & \rot{\bf FiQA-PL}         & \rot{\bf HotpotQA-PLHardNeg}   & \rot{\bf MSMARCO-PLHardNeg}   & \rot{\bf NFCorpus-PL}       & \rot{\bf NQ-PLHardNeg}   & \rot{\bf Quora-PLHardNeg}   & \rot{\bf SCIDOCS-PL}        & \rot{\bf SciFact-PL}       & \rot{\bf TRECCOVID-PL}      &  \bf Avg.           \\
\hline
 \multicolumn{8}{l}{\textbf{Small models (< 150M)}} \\
 \hline
 static-similarity-mrl-multilingual-v1        &          90 & 32.14             & 18.31                     & 7.54              & 24.62                      & 26.82                     & 17.17             & 12.23                & 65.41                   & 7.43              & 38.84            & 22.78             & 24.84             \\
 paraphrase-multilingual-MiniLM-L12-v2        &          81 & 37.86             & 22.34                     & 12.49             & 28.86                      & 38.43                     & 17.17             & 15.95                & 76.61                   & 10.26             & 40.23            & 34.22             & 30.40             \\
 multilingual-e5-small                        &          72 & 37.49             & 31.82                     & 22.02             & 61.51                      & 61.57                     & 26.50             & 42.09                & 77.70                   & 11.58             & 62.76            & 70.92             & 46.00             \\
 mmlw-e5-small                                &          72 & 54.21             & 35.39                     & 29.76             & 60.05                      & 54.73                     & 27.69             & 38.06                & 79.47                   & 14.90             & 58.41            & 58.09             & 46.43             \\
 st-polish-paraphrase-from-distilroberta      &         100 & 49.42             & 23.99                     & 19.57             & 29.26                      & 48.84                     & 22.52             & 23.52                & 80.08                   & 12.14             & 49.50            & 38.96             & 36.16             \\
 st-polish-paraphrase-from-mpnet              &         100 & 51.86             & 29.13                     & 22.28             & 36.27                      & 50.35                     & 24.04             & 26.12                & 80.61                   & 13.24             & 52.47            & 35.22             & 38.33             \\
 silver-retriever-base-v1.1                   &         100 & 47.07             & 31.69                     & 24.99             & 49.85                      & 62.15                     & 29.29             & 42.34                & 78.40                   & 11.04             & 52.80            & 42.53             & 42.92             \\
 mmlw-roberta-base                            &          90 & 59.04             & 40.33                     & 35.21             & 68.30                      & 64.07                     & 34.17             & 49.25                & 83.79                   & 17.95             & 66.00            & 71.48             & 53.60             \\
 distiluse-base-multilingual-cased-v2         &          81 & 36.70             & 17.48                     & 8.02              & 27.83                      & 27.58                     & 16.28             & 9.70                 & 71.46                   & 6.50              & 33.02            & 16.89             & 24.68             \\
  \hline
 \multicolumn{8}{l}{\textbf{Base models}} \\
 \hline
 drama-base                                   &          72 & 40.58             & 8.13                      & 11.49             & 29.35                      & 21.29                     & 21.35             & 3.65                 & 64.35                   & 11.22             & 58.05            & 41.73             & 28.29             \\
 paraphrase-multilingual-mpnet-base-v2        &          81 & 42.61             & 24.78                     & 14.71             & 34.08                      & 48.75                     & 18.54             & 17.23                & 77.81                   & 11.17             & 41.55            & 35.43             & 33.33             \\
 multilingual-e5-base                         &          72 & 42.86             & 31.94                     & 25.59             & 65.21                      & 64.64                     & 25.99             & 46.41                & 80.73                   & 12.36             & 62.27            & 65.90             & 47.63             \\
 mmlw-e5-base                                 &          72 & 58.45             & 41.17                     & 34.60             & 68.02                      & 64.20                     & 33.74             & 48.15                & 83.65                   & 17.39             & 68.31            & 73.07             & 53.70             \\
 snowflake-arctic-embed-m-v2.0                &          72 & 51.39             & 37.79                     & 33.38             & 67.41                      & 67.37                     & 30.57             & 45.61                & 80.94                   & 15.84             & 66.18            & 77.86             & 52.21             \\
 \hline
 \multicolumn{8}{l}{\textbf{Large models}} \\
 \hline
 drama-large                                  &          72 & 43.28             & 11.52                     & 16.11             & 34.38                      & 27.06                     & 24.06             & 6.10                 & 70.01                   & 12.24             & 62.01            & 58.64             & 33.22             \\
 mmlw-roberta-large                           &          90 & 63.66             & 21.46                     & 40.83             & 63.91                      & 58.54                     & 33.97             & 19.42                & \cellcolor{best} \textbf{86.05}          & 19.44             & 70.70            & 71.01             & 49.91             \\
 mmlw-retrieval-roberta-large                 &          81 & 58.73             & \cellcolor{second_best} \underline{44.81}         & 39.32             & 71.98                      & \cellcolor{second_best} \underline{74.21}         & 35.43             & 55.94                & 85.52                   & 18.57             & 72.41            & 72.65             & 57.23             \\
 mmlw-retrieval-roberta-large-v2              &         54 & 61.04             & 43.34                     & 44.91             & 68.99                      & 71.47                     & 37.48             & 59.81                & 82.05                   & 21.60             & 74.63            & 76.49             & 58.35             \\
 LaBSE                                        &         100 & 38.56             & 21.85                     & 7.66              & 28.82                      & 33.43                     & 17.45             & 14.04                & 73.79                   & 7.47              & 39.79            & 18.13             & 27.36             \\
 KaLM-embedding-multilingual-mini-instruct-v1 &          18 & 47.76             & 32.07                     & 24.50             & 61.30                      & 49.88                     & 27.12             & 32.72                & 74.12                   & 14.08             & 61.33            & 65.65             & 44.59             \\
 multilingual-e5-large                        &          72 & 52.99             & 36.52                     & 32.97             & 67.57                      & 70.79                     & 30.21             & 53.58                & 82.72                   & 13.82             & 65.66            & 69.86             & 52.43             \\
 mmlw-e5-large                                &          72 & 63.45             & 44.14                     & 39.99             & 72.10                      & 70.11                     & 34.12             & 50.66                & 85.06                   & 19.18             & 71.59            & 71.44             & 56.53             \\
 snowflake-arctic-embed-l-v2.0                &          81 & 54.61             & 39.73                     & 36.85             & 66.58                      & 69.58                     & 32.11             & 52.13                & 83.59                   & 17.04             & 67.94            & 76.98             & 54.29             \\
 Qwen3-Embedding-0.6B                         &          72 & 57.53             & 30.23                     & 27.38             & 58.31                      & 64.04                     & 26.83             & 34.29                & 79.02                   & 16.24             & 61.48            & 79.16             & 48.59             \\
 \hline
 \multicolumn{8}{l}{\textbf{Extra large models (>1B)}} \\
 \hline
 drama-1b                                     &          72 & 49.46             & 34.02                     & 35.13             & 68.41                      & 55.30                     & 33.01             & 34.54                & 81.62                   & 17.08             & 73.04            & 84.81             & 51.49             \\
 stella-pl                                    &          54 & 60.22             & \cellcolor{best} \textbf{46.2}             & \cellcolor{best} \textbf{52.03}    & 71.18                      & 72.62                     & \cellcolor{best} \textbf{39.94}    & \cellcolor{second_best} \underline{61.62}    & \cellcolor{second_best} \underline{85.67}       & \cellcolor{best} \textbf{23.54}    & \cellcolor{second_best} \underline{78.3} & 77.72             & \cellcolor{second_best} \underline{60.82} \\
 stella-pl-retrieval-8k                       &          54 & \cellcolor{second_best} \underline{66.03} & 44.36                     & \cellcolor{second_best} \underline{51.51} & \cellcolor{second_best} \underline{73.84}          & \cellcolor{best} \textbf{74.49}            & \cellcolor{second_best} \underline{39.56} & \cellcolor{best} \textbf{63.83}       & 85.07                   & 22.57             & \cellcolor{best} \textbf{79.67}   & 76.54             & \cellcolor{best} \textbf{61.59}    \\
 Qwen3-Embedding-4B                           &          72 & 64.14             & 39.16                     & 38.03             & 68.64                      & 70.05                     & 33.85             & 47.33                & 80.57                   & 20.97             & 72.81            & \cellcolor{second_best} \underline{87.61} & 56.65             \\
 Qwen3-Embedding-8B                           &          72 & \cellcolor{best} \textbf{66.82}    & 41.04                     & 44.53             & 70.48                      & 71.26                     & 35.45             & 50.53                & 82.34                   & \cellcolor{second_best} \underline{22.88} & 76.06            & \cellcolor{best} \textbf{89.93}    & 59.21             \\
 BGE-Multilingual-Gemma2                      &          54 & 59.24             & 43.67                     & 45.44             & \cellcolor{best} \textbf{74.73}             & 74.00                     & 36.89             & 57.42                & 84.08                   & 18.08             & 73.45            & 81.26             & 58.93             \\

 \hline
 \end{tabular}
\caption{\label{tab:retrieval-results}
Evaluation results on retrieval tasks using nDCG@10. The best scores for a given column are marked in \colorbox{best}{\textbf{bold}}, and the second best are \colorbox{second_best}{\underline{underlined}}. 
}
\end{table*}

\begin{table*}[h]
    \scriptsize
    \centering
\begin{tabular}{lr|llll}
\hline
 \bf Model name  &   \rot{\bf Zero shot} & \rot{\bf SICK-R-PL}      & \rot{\bf CDSC-R}         & \rot{\bf STSBenchmarkMultilingual}   & \bf Avg.           \\
\hline
 \multicolumn{6}{l}{\textbf{Small models (< 150M)}} \\
 \hline
 static-similarity-mrl-multilingual-v1        &         100 & 61.40             & 86.97             & 67.65                         & 72.01             \\
 multilingual-e5-small                        &         100 & 70.62             & 90.95             & 73.67                         & 78.41             \\
 paraphrase-multilingual-MiniLM-L12-v2        &         100 & 68.77             & 88.98             & 78.29                         & 78.68             \\
 mmlw-e5-small                                &         100 & 74.66             & 90.57             & 80.91                         & 82.05             \\
 silver-retriever-base-v1.1                   &         100 & 64.46             & 88.34             & 71.03                         & 74.61             \\
 st-polish-paraphrase-from-distilroberta      &         100 & 76.37             & 89.62             & 81.89                         & 82.63             \\
 st-polish-paraphrase-from-mpnet              &         100 & 76.18             & 88.56             & 83.75                         & 82.83             \\
 mmlw-roberta-base                            &         100 & 79.20             & 92.55             & 83.84                         & 85.20             \\
 distiluse-base-multilingual-cased-v2         &         100 & 65.53             & 87.67             & 74.06                         & 75.75             \\
  \hline
 \multicolumn{6}{l}{\textbf{Base models}} \\
 \hline
 drama-base                                   &         100 & 56.34             & 81.04             & 57.66                         & 65.01             \\
 mmlw-e5-base                                 &         100 & 43.11             & 59.57             & 44.39                         & 49.02             \\
 multilingual-e5-base                         &         100 & 71.46             & 89.61             & 76.32                         & 79.13             \\
 paraphrase-multilingual-mpnet-base-v2        &         100 & 73.13             & 88.80             & 81.46                         & 81.13             \\
 snowflake-arctic-embed-m-v2.0                &         100 & 66.57             & 90.22             & 70.00                         & 75.60             \\
 \hline
 \multicolumn{6}{l}{\textbf{Large models}} \\
 \hline
 drama-large                                  &         100 & 58.76             & 83.39             & 58.99                         & 67.05             \\
 mmlw-retrieval-roberta-large                 &         100 & 79.36             & \cellcolor{best} \textbf{92.78}    & 82.00                         & 84.71             \\
 mmlw-roberta-large                           &         100 & 79.91             & 92.54             & 83.25                         & 85.23             \\
 mmlw-retrieval-roberta-large-v2              &          66 & 80.90             & 91.68             & 84.35                         & 85.64             \\
 LaBSE                                        &         100 & 65.90             & 85.53             & 72.58                         & 74.67             \\
 KaLM-embedding-multilingual-mini-instruct-v1 &         100 & 66.58             & 90.00             & 72.13                         & 76.24             \\
 mmlw-e5-large                                &         100 & 33.98             & 40.00             & 45.86                         & 39.95             \\
 multilingual-e5-large                        &         100 & 74.86             & 89.80             & 79.57                         & 81.41             \\
 snowflake-arctic-embed-l-v2.0                &         100 & 68.86             & 90.38             & 74.61                         & 77.95             \\
 Qwen3-Embedding-0.6B                         &         100 & 69.63             & 88.32             & 77.40                         & 78.45             \\
 \hline
 \multicolumn{6}{l}{\textbf{Extra large models (>1B)}} \\
 \hline
 drama-1b                                     &         100 & 69.81             & 89.72             & 75.09                         & 78.21             \\
 stella-pl-retrieval-8k                       &          66 & \cellcolor{second_best} \underline{81.65} & 92.11             & 85.91                         & 86.56             \\
 stella-pl                                    &          66 & \cellcolor{best} \textbf{81.92}    & \cellcolor{second_best} \underline{92.68} & 86.02                         & \cellcolor{best} \textbf{86.87}    \\
 Qwen3-Embedding-4B                           &         100 & 77.85             & 91.43             & \cellcolor{second_best} \underline{87.37}             & 85.55             \\
 Qwen3-Embedding-8B                           &         100 & 80.11             & 91.60             & \cellcolor{best} \textbf{88.44}                & \cellcolor{second_best} \underline{86.72} \\
 BGE-Multilingual-Gemma2                      &         100 & 78.16             & 90.96             & 82.79                         & 83.97             \\
 \hline
 \end{tabular}
\caption{\label{tab:sts-results}
Evaluation results on STS tasks using Spearman correlation based on cosine similarity. The best scores for a given column are marked in \colorbox{best}{\textbf{bold}}, and the second best are \colorbox{second_best}{\underline{underlined}}. 
}
\end{table*}

\end{document}